\documentclass[11pt]{article}

\usepackage[preprint]{acl}

\usepackage{times}
\usepackage{latexsym}

\usepackage[T1]{fontenc}

\usepackage[utf8]{inputenc}

\usepackage{microtype}

\usepackage{inconsolata}

\usepackage{graphicx}

\usepackage{amsmath}
\usepackage{amssymb}
\usepackage{enumitem}
\usepackage{graphicx}
\usepackage{subcaption}
\usepackage{multirow}
\usepackage{booktabs}
\usepackage{hyperref}

\title{POP: Prefill-Only Pruning for Efficient Large Model Inference}

\author{
    \textbf{Junhui He\textsuperscript{1,2}} \quad
    \textbf{Zhihui Fu\textsuperscript{2}} \quad
    \textbf{Jun Wang\textsuperscript{2}} \quad
    \textbf{Qingan Li\textsuperscript{1} \thanks{Corresponding author.}} \quad
    \\
    \\
    \textsuperscript{1}Wuhan University \quad
    \textsuperscript{2}OPPO Research Institute
}

\begin{document}

\maketitle

\begin{abstract}
    Large Language Models (LLMs) and Vision-Language Models (VLMs) have demonstrated remarkable capabilities.
However, their deployment is hindered by significant computational costs. 
Existing structured pruning methods, while hardware-efficient, often suffer from significant accuracy degradation. 
In this paper, we argue that this failure stems from a stage-agnostic pruning approach that overlooks the asymmetric roles between the prefill and decode stages. 
By introducing a virtual gate mechanism, our importance analysis reveals that deep layers are critical for next-token prediction (decode) but largely redundant for context encoding (prefill). 
Leveraging this insight, we propose Prefill-Only Pruning (POP), a stage-aware inference strategy that safely omits deep layers during the computationally intensive prefill stage while retaining the full model for the sensitive decode stage. 
To enable the transition between stages, we introduce independent Key-Value (KV) projections to maintain cache integrity, and a boundary handling strategy to ensure the accuracy of the first generated token. 
Extensive experiments on Llama-3.1, Qwen3-VL, and Gemma-3 across diverse modalities demonstrate that POP achieves up to 1.37$\times$ speedup in prefill latency with minimal performance loss, effectively overcoming the accuracy-efficiency trade-off limitations of existing structured pruning methods.
\end{abstract}

\section{Introduction}

Large Language Models (LLMs) and Vision-Language Models (VLMs) have achieved remarkable success across various domains. 
However, their massive parameter counts impose substantial computational overhead during inference, limiting their deployment. 
To solve this challenge, model pruning has been explored as a means to remove redundant computation and accelerate inference.
While unstructured pruning methods~~\citep{sparsegpt, wanda} can preserve accuracy, they often require specialized hardware and kernels to realize speedups. 
Conversely, structured pruning methods~\citep{llm-pruner, slicegpt, shortgpt, laco, sleb}, which remove entire components like layers or channels, offer better hardware compatibility but often suffer from significant accuracy degradation,  particularly in open-ended generative tasks.

We argue that the failure of existing structured pruning methods stems from a stage-agnostic, ``one-size-fits-all'' approach that ignores the functional asymmetry of the inference process. 
Standard autoregressive inference consists of two distinct stages: prefill and decode. 
The prefill stage aims solely to encode the input history into Key-Value (KV) cache to provide context for future generation. 
In contrast, the decode stage has a dual role: it must encode the current token into the cache, while simultaneously modeling the probability distribution of the next token. 
Intuitively, these distinct roles imply different sensitivities to pruning, requiring an asymmetric pruning strategy.

Motivated by this intuition, we propose \textbf{Prefill-Only Pruning (POP)}, a novel strategy that accelerates the computationally intensive prefill stage while preserving the full model capacity for the sensitive decode stage. 
We first introduce the virtual gate mechanism for layer importance estimation, by approximating the loss increment on the calibration dataset when each layer is removed.
Then, we analyze the importance of layers during prefill and decode stage respectively (depicted in Section~\ref{sec:stage-aware-importance-analysis}, Figure~\ref{fig:motivation}), and uncover a striking disparity: deep layers are critical for the generation phase but are largely redundant for the context encoding phase. 
Leveraging this insight, we accelerate inference by pruning these deep layers exclusively during the prefill stage, while retaining the full model capacity for the decode stage. 
To ensure a seamless transition between the pruned and full stages, we further incorporate mechanisms to handle the missing KV states and the stage boundary.

Our main contributions are summarized as follows:
\begin{itemize}[nosep]
    \item We introduce a virtual gate mechanism to model the importance of each layer to the final loss, revealing the functional asymmetry of LLMs: deep layers are essential for decode but redundant for prefill.
    \item We propose Prefill-Only Pruning (POP), a stage-aware method that removes deep layers during prefill to reduce FLOPs, while retaining the full model for decode. 
    We employ independent KV projections to generate KV states for the pruned layers, and boundary handling to ensure the accuracy of the first generated token.
    \item We conduct extensive evaluations across diverse model families (Llama-3.1, Qwen3-VL, Gemma-3) and modalities. 
    Experimental results demonstrate that POP achieves significant speedups with minimal accuracy loss, effectively overcoming the limitations of stage-agnostic structured pruning.
\end{itemize}
\section{Preliminary}


\subsection{Transformer Inference and KV Cache}

We consider a standard decoder-only Transformer architecture~\citep{attention, llama-3, qwen3-vl, gemma-3}. 
Let $L$ denote the number of layers in the model. 
For a specific layer $l \in \{1, \dots, L\}$, let $x_l \in \mathbb{R}^{d}$ denote the input hidden state (where $d$ is the hidden dimension). 
The computation within layer $l$ typically consists of a Grouped-Query Attention (GQA) block or a Multi-Head Latent Attention (MLA) Block, followed by a Feed-Forward Network (FFN) block, both with residual connections and layer normalization.

The forward pass for the $l$-th layer can be expressed as:
\begin{equation}
    \begin{aligned}
    y_l & := x_l + \operatorname{Attn}(x_l, K_l^\text{past}, V_l^\text{past}) \\
    x_{l+1} & := y_l + \operatorname{FFN}(y_l)
    \end{aligned}
\end{equation}
where $K_l^\text{past}$ and $V_l^\text{past}$ represent the cached Keys and Values from previous tokens in the sequence.

\noindent \textbf{KV Cache Generation.} 
During the inference process, specifically for the attention mechanism, the model computes the Query ($q_l$), Key ($k_l$), and Value ($v_l$) for the current token using projection matrices $W_l^Q, W_l^K, W_l^V$. 
To capture positional information, Rotary Positional Embeddings (RoPE) are typically applied to the Queries and Keys. 
The computation for the new KV pairs of the current token is:
\begin{equation}
    \begin{aligned}
    k_l^\text{new} & := \operatorname{RoPE}(\operatorname{LN}(x_l)W_l^K) \\
    v_l^\text{new} & := \operatorname{LN}(x_l)W_l^V
    \end{aligned}
\end{equation}
where $\operatorname{LN}(\cdot)$ denotes the normalization layer.

To enable autoregressive generation without recomputing history, these new keys and values are appended to the cache:
\begin{equation}
    \begin{aligned}
    K_l^\text{current} & := \operatorname{Concat}(K_l^\text{past}, k_l^\text{new}) \\
    V_l^\text{current} & := \operatorname{Concat}(V_l^\text{past}, v_l^\text{new})
    \end{aligned}
\end{equation}
The attention output is then computed using the updated $K_l^\text{current}$ and $V_l^\text{current}$. 

\subsection{Layer Pruning Formulation}

Layer pruning aims to accelerate inference by removing entire layers—both the Attention and FFN blocks—while preserving the residual connections. 
Formally, let $S_\text{skip} \subset \{1, \dots, L\}$ be the set of indices representing the layers to be pruned. 
For any layer $l \in S_\text{skip}$, we bypass the computational blocks entirely. 
The propagation through a pruned layer is reduced to an identity mapping:
\begin{equation}
    \hat{x}_{l+1} := x_l, \quad \forall l \in S_\text{skip}
\end{equation}

In existing pruning approaches, the set $S_\text{skip}$ is applied in a stage-agnostic manner across both the prefill and decode stage. 
However, as we discuss in the following section, this approach ignores the asymmetrical functional goals of the two phases: the prefill phase focuses solely on context encoding, while the decoding phase focuses on both context encoding and next-token prediction.
\section{Method}


\subsection{Estimating Layer Importance with Virtual Gates}

To effectively identify and remove redundant computation, we first require a quantitative metric to measure the contribution of each layer to the model's overall performance.
Intuitively,  we define the importance score of the $l$-th layer as the increment of average loss on a calibration dataset when the layer is removed (pruned), while keeping other parameters unchanged.
We denote this importance score as $I_l$.

Calculating $I_l$ directly based on this definition by physically removing each layer and evaluating the model is computationally intensive, requiring $L$ separate inference passes for an $L$-layer model. 
To address this, we introduce \textbf{virtual gates}. 
We modify the forward pass of the $l$-th layer by multiplying the residual branches (Attention and FFN outputs) with a virtual scalar parameter $g_l$~\citep{importance-estimation}:
\begin{equation}
    \begin{aligned}
        \hat{y}_l & := x_l + \operatorname{Attn}(x_l, K_l^\text{past}, V_l^\text{past}) \odot g_l \\
        \hat{x}_{l+1} & := \hat{y}_l + \operatorname{FFN}(\hat{y}_l) \odot g_l
    \end{aligned}
\end{equation}
When $g_l = 1$, the layer functions identically to the original pre-trained model; when $g_l = 0$, the residual update is suppressed, and $\hat{x}_{l+1} = x_l$, effectively pruning the layer.

We estimate the importance score $I_l$ by approximating the change in loss $\mathcal{L}$ when $g_l$ shifts from 1 to 0, using a second-order Taylor expansion around $g_l=1$~\citep{obd, importance-estimation}:
\begin{equation}
    \begin{aligned}
        I_l & = \mathbb E \left[ \mathcal{L}_{g_l=0} - \mathcal{L}_{g_l=1} \right] \\
        & \approx \mathbb E \left[  \frac{\partial \mathcal{L}}{\partial g_l}(0-1) + \frac 1 2 \frac{\partial^2 \mathcal{L}}{\partial g_l^2} (0-1)^2 \right] \\
        & = \mathbb E \left[ \frac{\partial \mathcal{L}}{\partial g_l}\right] + \frac 1 2 \mathbb E \left[ \frac{\partial^2 \mathcal{L}}{\partial g_l^2} \right]
    \end{aligned}
\end{equation}

Calculating the second-order term $\frac{\partial^2 \mathcal L}{\partial g_l^2}$ directly is still computationally intensive.
To approximate this term efficiently, we leverage the properties of Fisher Information.
Specifically, we adopt a sampling-based strategy to satisfy the assumptions linking the Hessian to the gradient variance~\citep{limitations-of-efim}.
For each prompt $x$ in the calibration dataset, instead of using ground-truth targets, we sample the target response $\hat y \sim P_\theta(\cdot | x)$ from the model's distribution to compute the loss.
This approach aligns the data distribution with the model distribution, achieving two key simplifications for calculating $I_l$:

\noindent \textbf{Vanishing First-Order Term:}
Since the model minimizes loss on its own generated distribution, the expected first-order gradient is zero:
\begin{equation}
    \mathbb E\left[ \frac{\partial \mathcal L}{\partial g_l} \right] = 0
\end{equation}

\noindent \textbf{Hessian-Gradient Relation:}
Under this sampling strategy, the expected Hessian matches the second moment of the gradients (i.e., the Fisher Information Matrix):
\begin{equation}
    \mathbb{E} \left[ \frac{\partial^2 \mathcal{L}}{\partial g_l^2} \right] = \mathbb{E} \left[ \left( \frac{\partial \mathcal{L}}{\partial g_l} \right)^2 \right]
\end{equation}

By substituting these simplifications back into the Taylor expansion, we derive an efficient estimator that relies solely on the gradient computed during a single forward-backward pass on each calibration sample:
\begin{equation}
    \tilde I_l = \mathbb E \left[ \left( \frac{\partial \mathcal L}{\partial g_l} \right)^2 \right]
\end{equation}

By estimating layer importance with virtual gates, we  accurately capture the sensitivity of the model outputs to the pruning of specific layers, while avoiding the iterative removal of each layer, or the heavy computation of the second-order derivative.

 \subsection{Stage-Aware Importance Analysis}

 \label{sec:stage-aware-importance-analysis}

\begin{figure}[t]
    \centering
        \begin{subfigure}{\linewidth}
            \centering
            \includegraphics[width=0.85\textwidth]{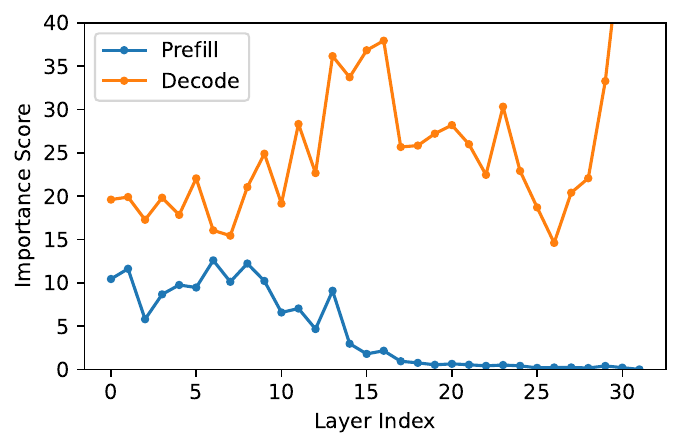}
            \caption{Llama-3.1-8B-Instruct, WizardLM-V2-196K}
        \end{subfigure} %
        \\
        \begin{subfigure}{\linewidth}
            \centering
            \includegraphics[width=0.85\textwidth]{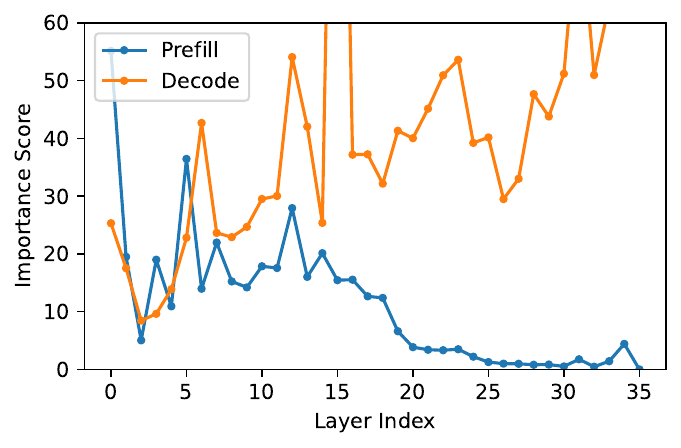}
            \caption{Qwen3-VL-8B-Instruct, LLAVA-Instruct-150K}
        \end{subfigure}
    \caption{Importance scores of layers from different models over datasets.}
    \label{fig:motivation}
\end{figure}

Consider the standard inference process of large models, which consists of two distinct stages: prefill and decode. 
The prefill stage has a singular role: to process the user's input prompt $x_{1:N-1}$ in parallel and encode the token information into the KV cache of every layer, providing context for future generation. 
In contrast, the decode stage processes the single latest token $x_t$ at each step. 
It must simultaneously fulfill a dual role: (1) encode the current token into the KV cache and append it to the sequence history; (2) model the probability distribution of the next token $x_{t+1}$ for autoregressive generation.

Despite sharing the same model parameters $\theta$, the functional roles of these two stages are asymmetric. 
Intuitively, these two stages might require different pruning strategies.
This motivates us to investigate the following questions:

\noindent \textit{\textbf{RQ1: Do prefill and decode stages exhibit asymmetric sensitivity to pruning?}}
Does one stage exhibit consistently higher sensitivity compared to the other, indicating greater fragility to pruning?

\noindent \textit{\textbf{RQ2: Do specific layers exhibit stage-dependent redundancy?}} 
Are there any layers that play critical roles in one stage, while being redundant in the other?

To answer these questions, we estimate the importance score of each layer during prefill and decode stage respectively, by extending the virtual gate mechanism to be stage-aware. 

Specifically, we treat the gates for the prefill and decode stages as separate parameters, denoted as $g^\text{prefill}$ and $g^\text{decode}$, respectively.
Let $\mathcal{E}$ and $\mathcal{D}$ represent the prefill and decoding processes.
The prefill stage takes the input prompt $x_{1:N-1}$ and outputs the KV cache for the context:
\begin{equation}
    Z_{1:N-1} = \{(K_l, V_l)\}_{l=1}^L := \mathcal{E}_{\theta, g^\text{prefill}}(x_{1:N-1})
\end{equation}
The decode stage takes the current token $x_t$, the generated history $x_{N:t-1}$, and the past KV cache $Z_{1:N-1}$ (from prefill) to predict the probability distribution of $x_{t+1}$:
\begin{equation}
    \begin{aligned}
        P_\theta(x_{t+1}|x_{1:t}) := \mathcal{D}_{\theta, g^\text{decode}}(x_{t+1} \mid & \mathcal{E}_{\theta, g^\text{prefill}}(x_{1:N-1}), \\
        & x_{N:t})
    \end{aligned}
\end{equation}

The final loss for the sequence is the cross-entropy over all output tokens:
\begin{equation}
    \begin{aligned}
        \mathcal{L} 
        = -\sum_{t=N}^{T-1} \log \mathcal{D}_{\theta, g^\text{decode}}(x_{t+1} \mid & \mathcal{E}_{\theta, g^\text{prefill}}(x_{1:N-1}), \\
        & x_{N:t})
    \end{aligned}
\end{equation}

We calculate the gradients for $g^\text{prefill}$ and $g^\text{decode}$ via a single forward-backward pass on each calibration sample, to obtain the stage-specific importance scores:
\begin{equation}
    \begin{aligned}
        \tilde{I}_l^\text{prefill} = \mathbb{E}[(\partial \mathcal{L} / \partial g_l^\text{prefill})^2] \\
        \quad \tilde{I}_l^\text{decode} = \mathbb{E}[(\partial \mathcal{L} / \partial g_l^\text{decode})^2]
    \end{aligned}
\end{equation}

We conducted experiments using Llama-3.1-8B-Instruct on the text dataset WizardLM-V2-196k, and Qwen3-VL-8B-Instruct on the multimodal dataset LLAVA-Instruct-150K. 
The calculated importance scores are visualized in Figure~\ref{fig:motivation}. 
From experimental results, we observe consistent characteristics across different models and modalities:

\noindent \textbf{Disparity between stages:} 
The importance scores for the prefill and decode stages are highly asymmetric. 
For a majority of layers, the decode importance (orange line) is significantly higher than the prefill importance (blue line), indicating that the decode stage is much more sensitive to model pruning.

\noindent \textbf{Criticality of Deep Layers for the Decode Stage:} 
For decode stage, deep layers are generally more important than shallow layers.
Specifically, for the orange line, the first few layers show moderate importance, while importance increases with depth. 
The final layers exhibit extremely high scores, often exceeding the visualization range, indicating criticality for next-token prediction.

\noindent \textbf{Redundancy of Deep Layers for the Prefill Stage:} 
The layer importance distribution of prefill stage is markedly different. 
For the blue line, the initial layers show moderate importance, indicating they are crucial for initial feature extraction. 
The intermediate layers show a decline in importance. 
Notably, the final layers exhibit low importance scores, often approaching zero, indicating redundancy for later generation.

These results validate our hypothesis: there is a disparity in overall sensitivity between stages; deep layers are essential for constructing the output distribution (decode) but are largely redundant for encoding the context information (prefill). 
These observations motivate us to propose \textbf{prefill-only pruning (POP), a stage-aware pruning method that improves prefill efficiency while preserving model accuracy}.

\subsection{Prefill-Only Pruning for Efficient Inference}

Based on the stage-aware importance profile in Section~\ref{sec:stage-aware-importance-analysis}, we adopt a static pruning strategy removing the deep layers in prefill stage, while retaining the full model for decode stage.
Specifically, we prune the last 1/3 of the layers, a ratio empirically selected to balance efficiency and accuracy.
Extensive experiments on the sensitivity to pruning ratio are presented in Section~\ref{sec:ablation}.

Implementation of this asymmetric strategy requires addressing two key challenges: missing KV caches for pruned layers, and the boundary handling between the prefill and decode stage.

\noindent \textbf{KV Cache Generation with Independent KV Projections.}
A naive skipping of layer $l$ during prefill would result in a missing KV cache ($K_l, V_l$). Since the decode stage uses the full model, it requires valid KV entries for all layers to perform attention over the input history.

To resolve this, we decouple the KV projection from the main computational block. For a pruned layer $l \in S_\text{skip}$ during prefill, we execute independent KV projections by: (1) Compute KV Cache: We apply the projection matrices to the input state $x_l$ to generate and store the cache:
\begin{equation}
    k_l := \operatorname{RoPE}(\operatorname{LN}(x_l)W^K), \quad v_l := \operatorname{LN}(x_l)W^V
\end{equation}
(2) Skip computation: We bypass the heavy Attention and FFN computations, passing the input directly to the next layer:
\begin{equation}
    \hat{x}_{l+1} := x_l
\end{equation}

Since the computational and memory access cost of the projections $(W^K, W^V)$ is negligible compared to the full Attention and FFN blocks ($<5\%$ for Llama-3, Qwen-3 and gemma-3 models), this method maintains the speedup benefits of pruning while ensuring the decoding stage has access to a complete KV cache.

A potential concern regarding independent KV projections is the representation mismatch for deep layers.
We provide further experiments and discussions in Appendix~\ref{sec:representation-mismatch} to address this concern.

\noindent \textbf{Boundary Handling for the Last Input Token.}
In a standard inference pipeline, the prefill stage processes tokens $x_{1:N}$ and predicts $x_{N+1}$. 
However, our analysis shows that deep layers are critical for next-token prediction. 
If we prune the deep layers when processing the last input token $x_N$, the accuracy of the first generated token will be degraded, leading to an accumulation of errors throughout the entire generation process.

To mitigate this, we redefine the boundary between stages. 
We define the \textbf{pruned prefill stage} as processing $x_{1:N-1}$. 
The processing of the last input token $x_N$ is treated as the \textbf{first decode step}.
This ensures that the prediction of the first new token utilizes the model's full capacity, while improving the efficiency of the computationally intensive prefill stage.
\begin{table*}[t]
    \centering
    \caption{
        \textbf{Accuracy comparison across different models and tasks.}
        "Avg" denotes the average score across all tasks. 
        The pruning ratios are indicated in parentheses.
        $\dagger$ denotes likelihood-based tasks; $\ddagger$ denotes open-ended generation tasks.
        \textbf{Bold} indicates the best results for all structured pruning methods.
        \textit{Italic} indicates unstructured pruning methods (Wanda).
    }
    \label{tab:main_results}
    \resizebox{\textwidth}{!}{
        \begin{tabular}{l|cccc|cc|cc|cccc|c}
        \toprule
        \multirow{2}{*}{\textbf{Method}} & \multicolumn{4}{c|}{\textbf{Common Sense$^\dagger$}} & \multicolumn{2}{c|}{\textbf{Math \& Code$^\ddagger$}} & \multicolumn{2}{c|}{\textbf{Long Context QA$^\ddagger$}} & \multicolumn{4}{c|}{\textbf{Multi-Modal$^\ddagger$}} & \multirow{2}{*}{\textbf{Avg}} \\
         & MMLU & HellaSwag & WinoG & PIQA & GSM8K & HumanEval & MultiFieldQA & HotpotQA & MMMU & RealWorldQA & TextVQA & ScreenSpot & \\
        \midrule
        \multicolumn{14}{c}{\textit{Llama-3.1-8B-Instruct}} \\
        \midrule
        Full Model & 68.33 & 79.50 & 74.40 & 81.12 & 79.68 & 68.29 & 54.57 & 55.66 & - & - & - & - & 70.19 \\
        \itshape Wanda (30\%) & \itshape 65.87 & \itshape 78.96 & \itshape 74.59 & \itshape 80.74 & \itshape 76.42 & \itshape 65.84 & \itshape 52.80 & \itshape 53.03 & \itshape - & \itshape - & \itshape - & \itshape - & \itshape 68.53\\
        SliceGPT (25\%) & 34.97 & 51.19 & 66.54 & 63.87 & 0.91 & 0.00 & 12.35 & 8.71 & - & - & - & - & 29.82 \\
        ShortGPT (25\%) & 65.80 & 61.93 & 69.77 & 70.51 & 0.38 & 0.00 & 6.80 & 3.81 & - & - & - & - & 34.88 \\
        \textbf{POP (31.25\%)} & \textbf{67.43} & \textbf{78.29} & \textbf{73.40} & \textbf{80.36} & \textbf{77.26} & \textbf{64.63} & \textbf{52.88} & \textbf{53.48} & - & - & - & - & \textbf{68.47} \\
        \midrule
        \multicolumn{14}{c}{\textit{Qwen3-VL-8B-Instruct}} \\
        \midrule
        Full Model & 74.95 & 76.60 & 73.72 & 79.92 & 81.50 & 92.07 & 53.53 & 65.49 & 51.33 & 69.67 & 82.24 & 87.03 & 74.00 \\
        \itshape Wanda (30\%) & \itshape \itshape \itshape 73.78 & \itshape \itshape \itshape 75.22 & \itshape \itshape \itshape 72.45 & \itshape \itshape \itshape 80.47 & \itshape \itshape \itshape 83.32 & \itshape \itshape \itshape 90.85 & \itshape \itshape 52.87 & \itshape \itshape 63.19 & \itshape 52.00 & \itshape 67.45 & \itshape 81.08 & \itshape 85.22 & \itshape 73.16 \\
        SliceGPT (25\%) & 39.16 & 44.50 & 57.93 & 67.25 & 13.95 & 17.68 & 40.76 & 38.33 & 28.00 & 32.55 & 13.54 & 0.24 & 32.82 \\
        ShortGPT (25\%) & 33.85 & 48.24 & 61.56 & 64.96 & 0.83 & 0.00 & 21.44 & 16.37 & 32.22 & 53.07 & 33.69 & 0.86 & 30.59 \\
        \textbf{POP (33.3\%)} & \textbf{75.05} & \textbf{76.44} & \textbf{73.88} & \textbf{80.14} & \textbf{80.21} & \textbf{89.63} & \textbf{52.34} & \textbf{63.13} & \textbf{50.67} & \textbf{69.28} & \textbf{80.73} & \textbf{86.40} & \textbf{73.16} \\
        \midrule
        \multicolumn{14}{c}{\textit{Gemma-3-12B-It}} \\
        \midrule
        Full Model & 71.46 & 81.96 & 74.35 & 78.07 & 73.62 & 82.32 & 55.90 & 59.62 & 46.78 & 54.64 & 67.02 & 11.08 & 63.07 \\
        \itshape Wanda (30\%) & \itshape69.70 & \itshape80.82 & \itshape73.64 & \itshape77.42 & \itshape 75.13 & \itshape 83.54 & \itshape55.28 & \itshape58.78 & \itshape45.89 & \itshape55.29 & \itshape 64.67 & \itshape10.38 & \itshape62.55 \\
        SliceGPT (25\%) & 22.95 & 34.12 & 54.14 & 55.93 & 1.67 & 0.00 & 10.83 & 4.18 & 25.56 & 5.23 & 2.59 & 0.24 & 18.12 \\
        ShortGPT (25\%) & 23.81 & 30.32 & 48.70 & 53.70 & 0.91 & 0.00 & 1.58 & 0.34 & 25.00 & 0.39 & 0.00 & 0.24 & 15.42 \\
        \textbf{POP (33.3\%)} & \textbf{71.37} & \textbf{81.96} & \textbf{74.59} & \textbf{79.76} & \textbf{73.16} & \textbf{81.10} & \textbf{57.33} & \textbf{59.11} & \textbf{46.78} & \textbf{55.42} & \textbf{63.71} & \textbf{11.08} & \textbf{62.95} \\
        \bottomrule
        \end{tabular}
    }
\end{table*}
\section{Experiments}

\subsection{Experimental Setup}

To comprehensively evaluate the effectiveness and generalization of our proposed POP, we conduct experiments across various model architectures, modalities, and downstream tasks. 

\noindent \textbf{Models.}
We select a diverse set of state-of-the-art open-weights models to demonstrate the generalization capability of our approach. 
Our selection covers both text-only models and vision-language models from different model series and sizes:
\begin{itemize}[nosep]
    \item \textbf{Text-Only Models:} We utilize Llama-3.1-8B-Instruct~\citep{llama-3} to evaluate performance on text understanding and generation tasks.
    \item \textbf{Vision-Language Models:} We utilize Qwen3-VL-8B-Instruct~\citep{qwen3-vl} and Gemma-3-12B-It~\citep{gemma-3} to evaluate performance on text understanding, text generation and vision understanding tasks.
\end{itemize}

\noindent \textbf{Methods.}
We compare POP with representative unstructured and structured pruning methods:
\begin{itemize}[nosep]
    \item \textbf{Unstructured}: We compare with Wanda~\citep{wanda}, an unstructured weight pruning method based on weight magnitudes and input activations.
    \item \textbf{Structured}: We compare with two methods: (1) SliceGPT~\citep{slicegpt}: removes rows and columns of weight matrices using PCA-based transformations;
    and (2) ShortGPT~\citep{shortgpt}: identifies and removes redundant layers based on cosine similarities of hidden states.
\end{itemize}

To ensure a fair comparison, we adjust the pruning ratio of all baselines to achieve a comparable FLOPs reduction during the prefill stage.

\noindent \textbf{Benchmarks.}
We employ a diverse set of benchmarks covering common sense reasoning, generative tasks, contextual understanding, and multimodal capabilities:
\begin{itemize}[nosep]
    \item \textbf{Common Sense:} We report 0-shot accuracy on MMLU~\citep{mmlu}, HellaSwag\citep{hellaswag}, Winogrande~\citep{winogrande}, and PIQA~\citep{piqa}.
    \item \textbf{Math \& Code:} We evaluate complex reasoning capabilities using GSM8K~\citep{gsm8k} for mathematics and HumanEval~\citep{humaneval} for code generation.
    \item \textbf{Long Context QA:} We evaluate long context understanding capabilities using MultiFieldQA for single-doc QA and HotpotQA for multi-doc QA~\citep{longbench}.
    \item \textbf{Multimodal Understanding:} For VLM evaluation, we use MMMU~\citep{mmmu} for multi-discipline understanding, RealWorldQA~\citep{realworldqa} for spatial reasoning, TextVQA~\citep{textvqa} for OCR-based QA, and ScreenSpot~\citep{screenspot} for GUI element localization.
\end{itemize}

For accuracy evaluations, we adopt two distinct strategies on different tasks. 
For common sense reasoning, we employ a likelihood-based approach: the model ranks candidate options based on their conditional probabilities (normalized by length), selecting the highest-scoring option as the prediction. 
Conversely, for other tasks, we utilize an open-ended generation approach: the model produces full responses via greedy decoding, which are then evaluated using exact match or functional correctness after rule-based answer extraction.

\noindent \textbf{Implementation Details.}
Calibration datasets for all methods consist of 200 samples from the WizardLM-V2-196K dataset~\citep{wizardlm-v2} for text-only models, or the LLAVA-Instruct-150K dataset~\citep{llava-instruct} for vision-language models.
All experiments are implemented in PyTorch~\citep{pytorch, pytorch-2.0} using the HuggingFace Transformers~\citep{transformers} library and executed on NVIDIA A100 80GB GPUs.
Evaluation on downstream tasks are conducted using the LM-Evaluation-Harness~\citep{lm-eval} library, the LongBench library~\citep{longbench} and the LMMs-Eval library~\citep{lmms-eval}.

\subsection{Accuracy on Downstream Tasks}

Table~\ref{tab:main_results} compares the pruning ratios and accuracies of different methods. Experimental results draw the following conclusions:

\noindent \textbf{Existing structured pruning methods exhibit catastrophic collapse on open-ended generation tasks.}
As shown in Table~\ref{tab:main_results}, while SliceGPT and ShortGPT maintain reasonable performance on likelihood-based tasks, they suffer from severe accuracy degradation on open-ended generation tasks. 
For instance, when applied to Llama-3.1, SliceGPT drops from 79.68\% to 0.91\% on GSM8K. 
Similarly, on the multimodal Qwen3-VL, SliceGPT degrades ScreenSpot accuracy from 87.03\% to 0.86\%. 
These results suggest that existing structured pruning methods destroy the generation capability of models.

\noindent \textbf{POP preserves model accuracies across benchmarks.}
In contrast, POP demonstrates remarkable stability across all task categories,  despite pruning a larger portion of the model ($\approx 33\%$) compared to the baselines ($\approx 25\%$).
More specifically, for generative reasoning tasks, POP achieves 77.26\% on GSM8K and 64.63\% on HumanEval when applied to Llama-3.1, retaining 97.00\% and 95.64\% of the full model's performance, respectively.
For long context QA tasks, POP also exhibits minimal performance drops (e.g., 59.11\% vs 59.62\% on HotpotQA when applied to Gemma-3).
The robustness extends to multimodal models and tasks. 
On Qwen3-VL, POP maintains near-lossless performance on MMMU (50.67\% vs 51.33\%) and ScreenSpot (86.40\% vs 87.03\%), significantly outperforming structured pruning baselines.

\noindent \textbf{POP achieves accuracies comparable to unstructured pruning methods, while offering better hardware compatibility.}
Wanda, being an unstructured pruning method, generally preserves accuracy better than traditional structured methods.
However, unstructured pruning methods requires specialized hardware and kernels for acceleration. 
POP achieves accuracy on par with Wanda across benchmarks (e.g., Gemma-3 Avg: 62.95\% vs 62.55\%) while offering much better hardware compatibility by structurally removing model layers.

\subsection{Inference Speedup}

We evaluate the inference speedup of POP by measuring the Time-to-First-Token (TTFT) on NVIDIA A100 GPUs. 
We conduct all experiments with a batch size of 8, utilizing text inputs with lengths ranging from 32 to 2048 tokens, and image inputs with resolutions ranging from $640 \times 480$ to $2560 \times 1440$.
Experimental results are shown in Table~\ref{tab:speedup_results}.

\noindent \textbf{Hardware Limitations for Unstructured Pruning.}
While Wanda achieves high accuracy on downstream tasks, it yields no wall-clock speedup (1.0$\times$) on our GPUs (A100) using dense kernels.
This result confirms that unstructured pruning theoretically reduces FLOPs but requires specialized hardware and sparse kernels to realize efficiency gains.

\noindent \textbf{Impact of Sequence Length for Text Inputs.}
For text inputs, we observe that the efficiency gains of POP are highly dependent on the input sequence length.
At short context lengths (e.g., 32 tokens), POP exhibits limited speedups (e.g., $1.22\times$ for Llama-3.1, $1.02 \times$ for Gemma-3).
This is primarily due to our boundary handling strategy. 
The short-input prefill is a memory-bound process, dominated by model weight access. 
Since processing the final input token requires using the full model, POP cannot reduce these memory access overheads, thus limiting performance gains.

However, as the sequence length increases, the computational cost of the first $N-1$ tokens (processed by the pruned model) becomes the dominant factor in TTFT.
Consequently, POP demonstrates significant speedup. 
At an input length of 2048, POP achieves a 1.36$\times$ speedup on Llama-3.1 and 1.37$\times$ on Gemma-3, outperforming both SliceGPT and ShortGPT. 
These results confirm that POP is particularly well-suited for compute-bound, long-context scenarios.

\noindent \textbf{Efficiency on Multimodal Tasks.}
For vision inputs, POP delivers speedups between 1.16$\times$ and 1.19$\times$, consistently surpassing SliceGPT and ShortGPT for all image resolutions, while offering much better accuracies.
These results confirm the advantage of POP in multimodal tasks.

Overall, experimental results validate that POP offers a practical "plug-and-play" acceleration solution that requires no model retraining or specialized hardware or kernels, making it particularly advantageous for long-context and high-resolution multimodal processing where prefill latency is critical.

\begin{table}[t]
    \centering
    \caption{\textbf{TTFT speedup comparison across different models and input lengths.} 
    All experiments are conducted with a batch size of 8.
    Values represent the speedup ratio relative to the full model ($1.0\times$).}
    \label{tab:speedup_results}
    \resizebox{\linewidth}{!}{
    \begin{tabular}{l|cccc}
        \toprule
        \multicolumn{5}{l}{\textbf{Llama-3.1-8B-Instruct}} \\
        \hline
        Method \textbackslash \ Input Length & 32 & 128 & 512 & 2048 \\
        \hline
        Wanda & 1.00 & 1.00 & 1.00 & 1.00 \\
        SliceGPT & 1.22 & \textbf{1.31} & 1.29 & 1.31 \\
        ShortGPT & \textbf{1.30} & 1.29 & 1.31 & 1.30 \\
        POP & 1.22 & 1.27 & \textbf{1.34} & \textbf{1.36} \\
        \midrule
        \multicolumn{5}{l}{\textbf{Gemma-3-12B-It}} \\
        \hline
        Method \textbackslash \ Input Length & 32 & 128 & 512 & 2048 \\
        \hline
        Wanda & 1.00 & 1.00 & 1.00 & 1.00 \\
        SliceGPT & 1.10 & \textbf{1.29} & 1.27 & 1.29 \\
        ShortGPT & \textbf{1.25} & \textbf{1.29} & 1.31 & 1.31 \\
        POP & 1.02 & 1.27 & \textbf{1.34} & \textbf{1.37} \\
        \midrule
        \multicolumn{5}{l}{\textbf{Qwen3-VL-8B-Instruct}} \\
        \hline
        Method \textbackslash \ Resolution & $640 \times480$ & $1280 \times 720$ & $1920 \times 1080$ & $2560\times1440$  \\
        \hline
        Wanda & 1.00 & 1.00 & 1.00 & 1.00 \\
        SliceGPT & 1.14 & 1.16 & 1.15 & 1.14 \\
        ShortGPT & 1.18 & 1.17 & 1.15 & 1.13 \\
        POP & \textbf{1.19} & \textbf{1.19} & \textbf{1.18} & \textbf{1.16} \\
        \bottomrule
    \end{tabular}
    }
\end{table}

\subsection{Ablation Study}
\label{sec:ablation}

To validate the design choices and parameter sensitivity of POP, we conduct comprehensive ablation studies using Qwen3-VL. 
We report the accuracy on GSM8K (complex reasoning) and HotpotQA (long-context understanding).

\subsubsection{Effectiveness of Design Choices}

We first verify the necessity of our three key design components: (1) targeting deep layers, (2) independent KV projections, and (3) boundary handling. 
Experimental results are shown in Table~\ref{tab:design-choices}.

\begin{table}[t]
    \centering
    \caption{
        \textbf{Ablation on design choices.} 
        We compare POP with different layer selection strategies and component removals on Qwen3-VL-8B-Instruct. 
        ``w/o Indep. KV'' denotes removing independent KV projections for pruned layers.
        ``w/o Boundary'' denotes removing the boundary handling for the last input token.
    }
    \label{tab:design-choices}
    \small{
    \begin{tabular}{l|cc}
        \toprule
        \textbf{Method Variants} & \textbf{GSM8K} & \textbf{HotpotQA} \\
        \midrule
        Full Model & 81.50 & 65.49 \\
        \textbf{POP} & \textbf{80.21} & \textbf{63.13} \\
        \midrule
        \multicolumn{3}{l}{\textit{Layer Selection Strategy}} \\
        Shallow Pruning & 0.15 & 0.00 \\
        Interleaved Pruning & 56.48 & 6.81 \\
        \midrule
        \multicolumn{3}{l}{\textit{Component Necessity}} \\
        w/o Indep. KV Proj. & 2.05 & 1.18 \\
        w/o Boundary Handling & 77.33 & 11.45 \\
        \bottomrule
    \end{tabular}
    }
\end{table}


\noindent \textbf{Layer Selection Strategy.} 
We compare POP against the shallow pruning (first 1/3 layers) and interleaved pruning (every 3\textsuperscript{rd} layer) strategies. 
    Both variants suffer from significant accuracy degradation, with shallow pruning dropping to nearly zero (0.15\% on GSM8K, 0\% on HotpotQA). 
These results validate that the redundancy in the prefill stage is non-uniform and specifically concentrated in the deep layers, whereas shallow layers remain critical.
    
\noindent \textbf{Necessity of Independent KV Projections.} 
We evaluate a variant that removes the independent KV projections, and skips the KV cache generation for pruned layers entirely.
With this variant, the model can only access the last input token and the generated tokens of the last 1/3 layers during the decode stage, while being unable to access initial input tokens.
This results in catastrophic collapse in model accuracy (2.05\% on GSM8K, 1.18\% on HotpotQA), confirming that while the residual updates of deep layers are redundant for prefill, their KV states are indispensable for the full model to perform attention computations during the decode stage.
    
\noindent \textbf{Importance of Boundary Handling.} 
We evaluate a variant that removes the boundary handling for the last input token. 
With this variant, the $x_N$ is also processed with the pruned prefill model. 
This variant suffers from obvious drop in accuracy on both tasks (80.21\% to 77.33\% on GSM8K, 63.13\% to 11.45\% on HotpotQA), 
indicating the necessity of processing the final token with the full model.

\subsubsection{Sensitivity to Pruning Ratio}

We further investigate the trade-off between inference efficiency and model performance by varying the pruning ratio from 20\% to 60\%. 
The prefill speedup is measured with a sequence length of 1024 and a batch size of 4.
Experimental Results are presented in Table~\ref{tab:ablation_ratio}.

We observe that at lower pruning ratios (20\%-25\%), the model maintains or even slightly surpasses the full model's accuracy (e.g., 83.09\% vs 81.50\% on GSM8K). 
We hypothesize that mild pruning may act as a regularization mechanism, filtering out noise in the deep layers.
However, these ratios offer limited speedup. 
Our default ratio of 33\% achieves considerable acceleration ($1.37\times$) with negligible accuracy loss.
Pushing the ratio beyond 50\% leads to a sharp decline in performance, particularly on HotpotQA, indicating that excessive pruning compromises the model's capacity to encode complex context information.

\begin{table}[t]
    \centering
    \caption{
        \textbf{Impact of pruning ratio.} 
        Performance and speedup trade-off at different pruning ratios on Qwen3-VL-8B-Instruct.
    }
    \label{tab:ablation_ratio}
    \small{
    \begin{tabular}{c|c|cc}
        \toprule
        \textbf{Pruning Ratio} & \textbf{Speedup} & \textbf{GSM8K} & \textbf{HotpotQA} \\
        \midrule
        0\% (Full Model) & 1.00$\times$ & 81.50 & 65.49 \\
        \midrule
        20\% & 1.19$\times$ & 83.09 & 65.46 \\
        25\% & 1.25$\times$ & 82.34 & 65.81 \\
        \textbf{33\% (Default)} & \textbf{1.37$\times$} & \textbf{80.21} & \textbf{63.13} \\
        40\% & 1.46$\times$ & 80.82 & 61.69 \\
        50\% & 1.67$\times$ & 78.54 & 34.69 \\
        60\% & 1.96$\times$ & 38.51 & 5.45 \\
        \bottomrule
    \end{tabular}
    }
\end{table}
\section{Related Work}

\label{sec:related-work}

\noindent \textbf{Model Pruning.}
Model pruning accelerates inference by removing redundant parameters. 
Unstructured pruning methods, such as SparseGPT~\citep{sparsegpt} and Wanda~\citep{wanda}, prune individual weights based on magnitude and activation norms. 
While preserving accuracy, they often require specialized kernels to achieve wall-clock speedup. 
Structured pruning addresses this by removing coarse-grained components like layers or channels. 
Component-wise methods such as LLM-Pruner~\citep{llm-pruner} and SliceGPT~\citep{slicegpt} employ dependency graphs or matrix factorizations to prune structural units. 
In contrast, layer-wise methods like ShortGPT~\citep{shortgpt}, LaCo~\citep{laco} and SLEB~\citep{sleb} demonstrate that specific layers in LLMs are redundant. 
However, existing structured pruning methods are typically stage-agnostic, applying the same reduced architecture across both prefill and decode stages. 
Our work challenges this paradigm by revealing that layer redundancy is highly stage-dependent, motivating a prefill-only pruning strategy.

\noindent \textbf{Token Pruning and Compression.}
Complementary to parameter reduction, token pruning accelerates inference by reducing the sequence length. 
For text inputs, perplexity-based methods such as LLMLingua~\citep{llm-lingua, llm-lingua-2} compress input length by selecting only the most informative tokens with a smaller model.
In contrast, attention-based methods such as PyramidInfer~\citep{pyramidinfer} and DAC~\citep{dac} determine token importance with attention weights.
In the multimodal domain, token pruning methods such as FastV~\citep{fastv} and DART~\citep{dart} mitigate the visual token redundancy by discarding or merging image tokens after several layers in the language model backbone, based on attention weights or token similarities.
These methods can be applied along with our proposed POP.

\noindent \textbf{Sparse Attention.}
Recent research also optimizes the attention mechanism to handle long contexts. 
For the compute-bound prefill stage, existing methods such as MInference~\citep{minference}, MMInference~\citep{mminference} and FlexPrefill~\citep{flexprefill} utilize block-sparse attention to bypass insignificant calculations. 
For the memory-bound decode stage, existing approaches such as Quest~\citep{quest}, PQCache~\citep{pqcache} and MagicPIG~\citep{magicpig} relieve the KV cache bottleneck by offloading KV cache to CPU memory, and perform sparse retrieval for computation. 
These methods can also be combined seamlessly with POP for further efficiency improvement.

\section{Conclusion}

In this work, we identify and exploit the asymmetric sensitivity to model pruning between the prefill and decode stages. 
Our analysis highlights that while deep layers are indispensable for generation (decode), they contribute minimally to context encoding (prefill). 
Based on this, we introduce POP, a simple yet effective strategy that accelerates the prefill stage by pruning deep layers, while preserving the full model for the decode stage.
By decoupling the computational pathways of context processing and token generation, POP  achieves prefill speedup of up to 1.37$\times$, while maintaining the accuracy comparable to the full model, significantly outperforming existing structured pruning methods. 
Our findings suggest that stage-aware optimization is a promising direction for efficient LLM inference, potentially extending beyond pruning to other techniques such as quantization and model architecture design.   
\section{Limitations}

While POP provides a compelling trade-off between efficiency and accuracy, we acknowledge several limitations.

First, unlike stage-agnostic pruning methods that permanently remove parameters to reduce memory footprint, POP requires the full model weights to be loaded for the decode stage. 
Consequently, it does not alleviate peak VRAM usage and is best suited for compute-bound rather than capacity-bound scenarios.

Second, our current implementation is based on a monolithic inference pipeline modified from the Transformers library. 
Recent advancements in inference systems such as DistServe~\citep{distserve} and Splitwise~\citep{splitwise} propose disaggregated systems that deploy prefill and decode instances on separate hardware resources. 
As POP naturally treats these two stages differently, it holds great potential for integration into these systems to further maximize cluster-level throughput. 
However, adapting POP for such distributed frameworks involves non-trivial engineering efforts, which we leave for future research.
\section*{Acknowledgments}

We thank all the reviewers for their insightful comments. 
This work is supported by the National Natural Science Foundation of China (No. 62472330).

\bibliography{custom}

@inproceedings{obd,
  author       = {Yann LeCun and
                  John S. Denker and
                  Sara A. Solla},
  editor       = {David S. Touretzky},
  title        = {Optimal Brain Damage},
  booktitle    = {Advances in Neural Information Processing Systems 2, {[NIPS} Conference,
                  Denver, Colorado, USA, November 27-30, 1989]},
  pages        = {598--605},
  publisher    = {Morgan Kaufmann},
  year         = {1989},
  url          = {http://papers.nips.cc/paper/250-optimal-brain-damage},
  bibsource    = {dblp computer science bibliography, https://dblp.org}
}

@inproceedings{sparsegpt,
  author       = {Elias Frantar and
                  Dan Alistarh},
  editor       = {Andreas Krause and
                  Emma Brunskill and
                  Kyunghyun Cho and
                  Barbara Engelhardt and
                  Sivan Sabato and
                  Jonathan Scarlett},
  title        = {SparseGPT: Massive Language Models Can be Accurately Pruned in One-Shot},
  booktitle    = {International Conference on Machine Learning, {ICML} 2023, 23-29 July
                  2023, Honolulu, Hawaii, {USA}},
  series       = {Proceedings of Machine Learning Research},
  volume       = {202},
  pages        = {10323--10337},
  publisher    = {{PMLR}},
  year         = {2023},
  url          = {https://proceedings.mlr.press/v202/frantar23a.html},
  bibsource    = {dblp computer science bibliography, https://dblp.org}
}

@inproceedings{wanda,
  author       = {Mingjie Sun and
                  Zhuang Liu and
                  Anna Bair and
                  J. Zico Kolter},
  title        = {A Simple and Effective Pruning Approach for Large Language Models},
  booktitle    = {The Twelfth International Conference on Learning Representations,
                  {ICLR} 2024, Vienna, Austria, May 7-11, 2024},
  publisher    = {OpenReview.net},
  year         = {2024},
  url          = {https://openreview.net/forum?id=PxoFut3dWW},
  bibsource    = {dblp computer science bibliography, https://dblp.org}
}

@inproceedings{llm-pruner,
  author       = {Xinyin Ma and
                  Gongfan Fang and
                  Xinchao Wang},
  editor       = {Alice Oh and
                  Tristan Naumann and
                  Amir Globerson and
                  Kate Saenko and
                  Moritz Hardt and
                  Sergey Levine},
  title        = {LLM-Pruner: On the Structural Pruning of Large Language Models},
  booktitle    = {Advances in Neural Information Processing Systems 36: Annual Conference
                  on Neural Information Processing Systems 2023, NeurIPS 2023, New Orleans,
                  LA, USA, December 10 - 16, 2023},
  year         = {2023},
  url          = {http://papers.nips.cc/paper\_files/paper/2023/hash/44956951349095f74492a5471128a7e0-Abstract-Conference.html},
  bibsource    = {dblp computer science bibliography, https://dblp.org}
}

@inproceedings{slicegpt,
  author       = {Saleh Ashkboos and
                  Maximilian L. Croci and
                  Marcelo Gennari Do Nascimento and
                  Torsten Hoefler and
                  James Hensman},
  title        = {SliceGPT: Compress Large Language Models by Deleting Rows and Columns},
  booktitle    = {The Twelfth International Conference on Learning Representations,
                  {ICLR} 2024, Vienna, Austria, May 7-11, 2024},
  publisher    = {OpenReview.net},
  year         = {2024},
  url          = {https://openreview.net/forum?id=vXxardq6db},
  bibsource    = {dblp computer science bibliography, https://dblp.org}
}

@inproceedings{shortgpt,
  author       = {Xin Men and
                  Mingyu Xu and
                  Qingyu Zhang and
                  Qianhao Yuan and
                  Bingning Wang and
                  Hongyu Lin and
                  Yaojie Lu and
                  Xianpei Han and
                  Weipeng Chen},
  editor       = {Wanxiang Che and
                  Joyce Nabende and
                  Ekaterina Shutova and
                  Mohammad Taher Pilehvar},
  title        = {ShortGPT: Layers in Large Language Models are More Redundant Than
                  You Expect},
  booktitle    = {Findings of the Association for Computational Linguistics, {ACL} 2025,
                  Vienna, Austria, July 27 - August 1, 2025},
  pages        = {20192--20204},
  publisher    = {Association for Computational Linguistics},
  year         = {2025},
  url          = {https://aclanthology.org/2025.findings-acl.1035/},
  bibsource    = {dblp computer science bibliography, https://dblp.org}
}

@inproceedings{laco,
  author       = {Yifei Yang and
                  Zouying Cao and
                  Hai Zhao},
  editor       = {Yaser Al{-}Onaizan and
                  Mohit Bansal and
                  Yun{-}Nung Chen},
  title        = {LaCo: Large Language Model Pruning via Layer Collapse},
  booktitle    = {Findings of the Association for Computational Linguistics: {EMNLP}
                  2024, Miami, Florida, USA, November 12-16, 2024},
  pages        = {6401--6417},
  publisher    = {Association for Computational Linguistics},
  year         = {2024},
  url          = {https://doi.org/10.18653/v1/2024.findings-emnlp.372},
  doi          = {10.18653/V1/2024.FINDINGS-EMNLP.372},
  bibsource    = {dblp computer science bibliography, https://dblp.org}
}

@inproceedings{sleb,
  author       = {Jiwon Song and
                  Kyungseok Oh and
                  Taesu Kim and
                  Hyungjun Kim and
                  Yulhwa Kim and
                  Jae{-}Joon Kim},
  title        = {{SLEB:} Streamlining LLMs through Redundancy Verification and Elimination
                  of Transformer Blocks},
  booktitle    = {Forty-first International Conference on Machine Learning, {ICML} 2024,
                  Vienna, Austria, July 21-27, 2024},
  publisher    = {OpenReview.net},
  year         = {2024},
  url          = {https://openreview.net/forum?id=fuX4hyLPmO},
  bibsource    = {dblp computer science bibliography, https://dblp.org}
}

@article{llama-3,
  author       = {Llama Team},
  title        = {The Llama 3 Herd of Models},
  journal      = {CoRR},
  volume       = {abs/2407.21783},
  year         = {2024},
  url          = {https://doi.org/10.48550/arXiv.2407.21783},
  doi          = {10.48550/ARXIV.2407.21783},
  eprinttype    = {arXiv},
  eprint       = {2407.21783},
  bibsource    = {dblp computer science bibliography, https://dblp.org}
}

@misc{qwen3-vl,
      title={Qwen3-VL Technical Report}, 
      author={Shuai Bai and Yuxuan Cai and Ruizhe Chen and Keqin Chen and Xionghui Chen and Zesen Cheng and Lianghao Deng and Wei Ding and Chang Gao and Chunjiang Ge and Wenbin Ge and Zhifang Guo and Qidong Huang and Jie Huang and Fei Huang and Binyuan Hui and Shutong Jiang and Zhaohai Li and Mingsheng Li and Mei Li and Kaixin Li and Zicheng Lin and Junyang Lin and Xuejing Liu and Jiawei Liu and Chenglong Liu and Yang Liu and Dayiheng Liu and Shixuan Liu and Dunjie Lu and Ruilin Luo and Chenxu Lv and Rui Men and Lingchen Meng and Xuancheng Ren and Xingzhang Ren and Sibo Song and Yuchong Sun and Jun Tang and Jianhong Tu and Jianqiang Wan and Peng Wang and Pengfei Wang and Qiuyue Wang and Yuxuan Wang and Tianbao Xie and Yiheng Xu and Haiyang Xu and Jin Xu and Zhibo Yang and Mingkun Yang and Jianxin Yang and An Yang and Bowen Yu and Fei Zhang and Hang Zhang and Xi Zhang and Bo Zheng and Humen Zhong and Jingren Zhou and Fan Zhou and Jing Zhou and Yuanzhi Zhu and Ke Zhu},
      year={2025},
      eprint={2511.21631},
      archivePrefix={arXiv},
      primaryClass={cs.CV},
      url={https://arxiv.org/abs/2511.21631}, 
}

@article{gemma-3,
  author       = {Gemma Team},
  title        = {Gemma 3 Technical Report},
  journal      = {CoRR},
  volume       = {abs/2503.19786},
  year         = {2025},
  url          = {https://doi.org/10.48550/arXiv.2503.19786},
  doi          = {10.48550/ARXIV.2503.19786},
  eprinttype    = {arXiv},
  eprint       = {2503.19786},
  bibsource    = {dblp computer science bibliography, https://dblp.org}
}

@inproceedings{importance-estimation,
  author       = {Pavlo Molchanov and
                  Arun Mallya and
                  Stephen Tyree and
                  Iuri Frosio and
                  Jan Kautz},
  title        = {Importance Estimation for Neural Network Pruning},
  booktitle    = {{IEEE} Conference on Computer Vision and Pattern Recognition, {CVPR}
                  2019, Long Beach, CA, USA, June 16-20, 2019},
  pages        = {11264--11272},
  publisher    = {Computer Vision Foundation / {IEEE}},
  year         = {2019},
  url          = {http://openaccess.thecvf.com/content\_CVPR\_2019/html/Molchanov\_Importance\_Estimation\_for\_Neural\_Network\_Pruning\_CVPR\_2019\_paper.html},
  doi          = {10.1109/CVPR.2019.01152},
  bibsource    = {dblp computer science bibliography, https://dblp.org}
}

@inproceedings{mmlu,
  author       = {Dan Hendrycks and
                  Collin Burns and
                  Steven Basart and
                  Andy Zou and
                  Mantas Mazeika and
                  Dawn Song and
                  Jacob Steinhardt},
  title        = {Measuring Massive Multitask Language Understanding},
  booktitle    = {9th International Conference on Learning Representations, {ICLR} 2021,
                  Virtual Event, Austria, May 3-7, 2021},
  publisher    = {OpenReview.net},
  year         = {2021},
  url          = {https://openreview.net/forum?id=d7KBjmI3GmQ},
  bibsource    = {dblp computer science bibliography, https://dblp.org}
}

@inproceedings{hellaswag,
  author       = {Rowan Zellers and
                  Ari Holtzman and
                  Yonatan Bisk and
                  Ali Farhadi and
                  Yejin Choi},
  editor       = {Anna Korhonen and
                  David R. Traum and
                  Llu{\'{\i}}s M{\`{a}}rquez},
  title        = {HellaSwag: Can a Machine Really Finish Your Sentence?},
  booktitle    = {Proceedings of the 57th Conference of the Association for Computational
                  Linguistics, {ACL} 2019, Florence, Italy, July 28- August 2, 2019,
                  Volume 1: Long Papers},
  pages        = {4791--4800},
  publisher    = {Association for Computational Linguistics},
  year         = {2019},
  url          = {https://doi.org/10.18653/v1/p19-1472},
  doi          = {10.18653/V1/P19-1472},
  bibsource    = {dblp computer science bibliography, https://dblp.org}
}

@inproceedings{piqa,
  author       = {Yonatan Bisk and
                  Rowan Zellers and
                  Ronan Le Bras and
                  Jianfeng Gao and
                  Yejin Choi},
  title        = {{PIQA:} Reasoning about Physical Commonsense in Natural Language},
  booktitle    = {The Thirty-Fourth {AAAI} Conference on Artificial Intelligence, {AAAI}
                  2020, The Thirty-Second Innovative Applications of Artificial Intelligence
                  Conference, {IAAI} 2020, The Tenth {AAAI} Symposium on Educational
                  Advances in Artificial Intelligence, {EAAI} 2020, New York, NY, USA,
                  February 7-12, 2020},
  pages        = {7432--7439},
  publisher    = {{AAAI} Press},
  year         = {2020},
  url          = {https://doi.org/10.1609/aaai.v34i05.6239},
  doi          = {10.1609/AAAI.V34I05.6239},
  bibsource    = {dblp computer science bibliography, https://dblp.org}
}

@inproceedings{winogrande,
  author       = {Keisuke Sakaguchi and
                  Ronan Le Bras and
                  Chandra Bhagavatula and
                  Yejin Choi},
  title        = {WinoGrande: An Adversarial Winograd Schema Challenge at Scale},
  booktitle    = {The Thirty-Fourth {AAAI} Conference on Artificial Intelligence, {AAAI}
                  2020, The Thirty-Second Innovative Applications of Artificial Intelligence
                  Conference, {IAAI} 2020, The Tenth {AAAI} Symposium on Educational
                  Advances in Artificial Intelligence, {EAAI} 2020, New York, NY, USA,
                  February 7-12, 2020},
  pages        = {8732--8740},
  publisher    = {{AAAI} Press},
  year         = {2020},
  url          = {https://doi.org/10.1609/aaai.v34i05.6399},
  doi          = {10.1609/AAAI.V34I05.6399},
  bibsource    = {dblp computer science bibliography, https://dblp.org}
}

@inproceedings{longbench,
  author       = {Yushi Bai and
                  Xin Lv and
                  Jiajie Zhang and
                  Hongchang Lyu and
                  Jiankai Tang and
                  Zhidian Huang and
                  Zhengxiao Du and
                  Xiao Liu and
                  Aohan Zeng and
                  Lei Hou and
                  Yuxiao Dong and
                  Jie Tang and
                  Juanzi Li},
  editor       = {Lun{-}Wei Ku and
                  Andre Martins and
                  Vivek Srikumar},
  title        = {LongBench: {A} Bilingual, Multitask Benchmark for Long Context Understanding},
  booktitle    = {Proceedings of the 62nd Annual Meeting of the Association for Computational
                  Linguistics (Volume 1: Long Papers), {ACL} 2024, Bangkok, Thailand,
                  August 11-16, 2024},
  pages        = {3119--3137},
  publisher    = {Association for Computational Linguistics},
  year         = {2024},
  url          = {https://doi.org/10.18653/v1/2024.acl-long.172},
  doi          = {10.18653/V1/2024.ACL-LONG.172},
  bibsource    = {dblp computer science bibliography, https://dblp.org}
}

@inproceedings{mmmu,
  author       = {Xiang Yue and
                  Yuansheng Ni and
                  Tianyu Zheng and
                  Kai Zhang and
                  Ruoqi Liu and
                  Ge Zhang and
                  Samuel Stevens and
                  Dongfu Jiang and
                  Weiming Ren and
                  Yuxuan Sun and
                  Cong Wei and
                  Botao Yu and
                  Ruibin Yuan and
                  Renliang Sun and
                  Ming Yin and
                  Boyuan Zheng and
                  Zhenzhu Yang and
                  Yibo Liu and
                  Wenhao Huang and
                  Huan Sun and
                  Yu Su and
                  Wenhu Chen},
  title        = {{MMMU:} {A} Massive Multi-Discipline Multimodal Understanding and
                  Reasoning Benchmark for Expert {AGI}},
  booktitle    = {{IEEE/CVF} Conference on Computer Vision and Pattern Recognition,
                  {CVPR} 2024, Seattle, WA, USA, June 16-22, 2024},
  pages        = {9556--9567},
  publisher    = {{IEEE}},
  year         = {2024},
  url          = {https://doi.org/10.1109/CVPR52733.2024.00913},
  doi          = {10.1109/CVPR52733.2024.00913},
  bibsource    = {dblp computer science bibliography, https://dblp.org}
}

@misc{realworldqa,
  author       = {xAI},
  title        = {RealWorldQA Dataset},
  year         = {2024},
  publisher    = {Hugging Face},
  howpublished = {\url{https://huggingface.co/datasets/xai-org/RealworldQA}},
  note         = {Accessed: 2026-01-04}
}

@inproceedings{textvqa,
  author       = {Amanpreet Singh and
                  Vivek Natarajan and
                  Meet Shah and
                  Yu Jiang and
                  Xinlei Chen and
                  Dhruv Batra and
                  Devi Parikh and
                  Marcus Rohrbach},
  title        = {Towards {VQA} Models That Can Read},
  booktitle    = {{IEEE} Conference on Computer Vision and Pattern Recognition, {CVPR}
                  2019, Long Beach, CA, USA, June 16-20, 2019},
  pages        = {8317--8326},
  publisher    = {Computer Vision Foundation / {IEEE}},
  year         = {2019},
  url          = {http://openaccess.thecvf.com/content\_CVPR\_2019/html/Singh\_Towards\_VQA\_Models\_That\_Can\_Read\_CVPR\_2019\_paper.html},
  doi          = {10.1109/CVPR.2019.00851},
  bibsource    = {dblp computer science bibliography, https://dblp.org}
}

@inproceedings{screenspot,
  author       = {Kanzhi Cheng and
                  Qiushi Sun and
                  Yougang Chu and
                  Fangzhi Xu and
                  Yantao Li and
                  Jianbing Zhang and
                  Zhiyong Wu},
  editor       = {Lun{-}Wei Ku and
                  Andre Martins and
                  Vivek Srikumar},
  title        = {SeeClick: Harnessing {GUI} Grounding for Advanced Visual {GUI} Agents},
  booktitle    = {Proceedings of the 62nd Annual Meeting of the Association for Computational
                  Linguistics (Volume 1: Long Papers), {ACL} 2024, Bangkok, Thailand,
                  August 11-16, 2024},
  pages        = {9313--9332},
  publisher    = {Association for Computational Linguistics},
  year         = {2024},
  url          = {https://doi.org/10.18653/v1/2024.acl-long.505},
  doi          = {10.18653/V1/2024.ACL-LONG.505},
  bibsource    = {dblp computer science bibliography, https://dblp.org}
}

@article{gsm8k,
  title={Training Verifiers to Solve Math Word Problems},
  author={Cobbe, Karl and Kosaraju, Vineet and Bavarian, Mohammad and Chen, Mark and Jun, Heewoo and Kaiser, Lukasz and Plappert, Matthias and Tworek, Jerry and Hilton, Jacob and Nakano, Reiichiro and Hesse, Christopher and Schulman, John},
  journal={arXiv preprint arXiv:2110.14168},
  year={2021}
}

@misc{humaneval,
      title={Evaluating Large Language Models Trained on Code},
      author={Mark Chen and Jerry Tworek and Heewoo Jun and Qiming Yuan and Henrique Ponde de Oliveira Pinto and Jared Kaplan and Harri Edwards and Yuri Burda and Nicholas Joseph and Greg Brockman and Alex Ray and Raul Puri and Gretchen Krueger and Michael Petrov and Heidy Khlaaf and Girish Sastry and Pamela Mishkin and Brooke Chan and Scott Gray and Nick Ryder and Mikhail Pavlov and Alethea Power and Lukasz Kaiser and Mohammad Bavarian and Clemens Winter and Philippe Tillet and Felipe Petroski Such and Dave Cummings and Matthias Plappert and Fotios Chantzis and Elizabeth Barnes and Ariel Herbert-Voss and William Hebgen Guss and Alex Nichol and Alex Paino and Nikolas Tezak and Jie Tang and Igor Babuschkin and Suchir Balaji and Shantanu Jain and William Saunders and Christopher Hesse and Andrew N. Carr and Jan Leike and Josh Achiam and Vedant Misra and Evan Morikawa and Alec Radford and Matthew Knight and Miles Brundage and Mira Murati and Katie Mayer and Peter Welinder and Bob McGrew and Dario Amodei and Sam McCandlish and Ilya Sutskever and Wojciech Zaremba},
      year={2021},
      eprint={2107.03374},
      archivePrefix={arXiv},
      primaryClass={cs.LG}
}

@inproceedings{wizardlm-v2,
  author       = {Can Xu and
                  Qingfeng Sun and
                  Kai Zheng and
                  Xiubo Geng and
                  Pu Zhao and
                  Jiazhan Feng and
                  Chongyang Tao and
                  Qingwei Lin and
                  Daxin Jiang},
  title        = {WizardLM: Empowering Large Pre-Trained Language Models to Follow Complex
                  Instructions},
  booktitle    = {The Twelfth International Conference on Learning Representations,
                  {ICLR} 2024, Vienna, Austria, May 7-11, 2024},
  publisher    = {OpenReview.net},
  year         = {2024},
  url          = {https://openreview.net/forum?id=CfXh93NDgH},
  bibsource    = {dblp computer science bibliography, https://dblp.org}
}

@inproceedings{llava-instruct,
  author       = {Haotian Liu and
                  Chunyuan Li and
                  Qingyang Wu and
                  Yong Jae Lee},
  editor       = {Alice Oh and
                  Tristan Naumann and
                  Amir Globerson and
                  Kate Saenko and
                  Moritz Hardt and
                  Sergey Levine},
  title        = {Visual Instruction Tuning},
  booktitle    = {Advances in Neural Information Processing Systems 36: Annual Conference
                  on Neural Information Processing Systems 2023, NeurIPS 2023, New Orleans,
                  LA, USA, December 10 - 16, 2023},
  year         = {2023},
  url          = {http://papers.nips.cc/paper\_files/paper/2023/hash/6dcf277ea32ce3288914faf369fe6de0-Abstract-Conference.html},
  bibsource    = {dblp computer science bibliography, https://dblp.org}
}

@inproceedings{pytorch-2.0,
  author       = {Peng Wu},
  editor       = {Christophe Dubach and
                  Derek Bruening and
                  Ben Hardekopf},
  title        = {PyTorch 2.0: The Journey to Bringing Compiler Technologies to the
                  Core of PyTorch (Keynote)},
  booktitle    = {Proceedings of the 21st {ACM/IEEE} International Symposium on Code
                  Generation and Optimization, {CGO} 2023, Montr{\'{e}}al, QC,
                  Canada, 25 February 2023- 1 March 2023},
  pages        = {1},
  publisher    = {{ACM}},
  year         = {2023},
  url          = {https://doi.org/10.1145/3579990.3583093},
  doi          = {10.1145/3579990.3583093},
  bibsource    = {dblp computer science bibliography, https://dblp.org}
}

@article{transformers,
  author       = {Thomas Wolf and
                  Lysandre Debut and
                  Victor Sanh and
                  Julien Chaumond and
                  Clement Delangue and
                  Anthony Moi and
                  Pierric Cistac and
                  Tim Rault and
                  R{\'{e}}mi Louf and
                  Morgan Funtowicz and
                  Jamie Brew},
  title        = {HuggingFace's Transformers: State-of-the-art Natural Language Processing},
  journal      = {CoRR},
  volume       = {abs/1910.03771},
  year         = {2019},
  url          = {http://arxiv.org/abs/1910.03771},
  eprinttype    = {arXiv},
  eprint       = {1910.03771},
  bibsource    = {dblp computer science bibliography, https://dblp.org}
}

@misc{lm-eval,
  author       = {Gao, Leo and Tow, Jonathan and Abbasi, Baber and Biderman, Stella and Black, Sid and DiPofi, Anthony and Foster, Charles and Golding, Laurence and Hsu, Jeffrey and Le Noac'h, Alain and Li, Haonan and McDonell, Kyle and Muennighoff, Niklas and Ociepa, Chris and Phang, Jason and Reynolds, Laria and Schoelkopf, Hailey and Skowron, Aviya and Sutawika, Lintang and Tang, Eric and Thite, Anish and Wang, Ben and Wang, Kevin and Zou, Andy},
  title        = {The Language Model Evaluation Harness},
  month        = 07,
  year         = 2024,
  publisher    = {Zenodo},
  version      = {v0.4.3},
  doi          = {10.5281/zenodo.12608602},
  url          = {https://zenodo.org/records/12608602}
}

@inproceedings{lmms-eval,
  author       = {Kaichen Zhang and
                  Bo Li and
                  Peiyuan Zhang and
                  Fanyi Pu and
                  Joshua Adrian Cahyono and
                  Kairui Hu and
                  Shuai Liu and
                  Yuanhan Zhang and
                  Jingkang Yang and
                  Chunyuan Li and
                  Ziwei Liu},
  editor       = {Luis Chiruzzo and
                  Alan Ritter and
                  Lu Wang},
  title        = {LMMs-Eval: Reality Check on the Evaluation of Large Multimodal Models},
  booktitle    = {Findings of the Association for Computational Linguistics: {NAACL}
                  2025, Albuquerque, New Mexico, USA, April 29 - May 4, 2025},
  pages        = {881--916},
  publisher    = {Association for Computational Linguistics},
  year         = {2025},
  url          = {https://doi.org/10.18653/v1/2025.findings-naacl.51},
  doi          = {10.18653/V1/2025.FINDINGS-NAACL.51},
  bibsource    = {dblp computer science bibliography, https://dblp.org}
}

@inproceedings{llm-lingua,
  author       = {Huiqiang Jiang and
                  Qianhui Wu and
                  Chin{-}Yew Lin and
                  Yuqing Yang and
                  Lili Qiu},
  editor       = {Houda Bouamor and
                  Juan Pino and
                  Kalika Bali},
  title        = {LLMLingua: Compressing Prompts for Accelerated Inference of Large
                  Language Models},
  booktitle    = {Proceedings of the 2023 Conference on Empirical Methods in Natural
                  Language Processing, {EMNLP} 2023, Singapore, December 6-10, 2023},
  pages        = {13358--13376},
  publisher    = {Association for Computational Linguistics},
  year         = {2023},
  url          = {https://doi.org/10.18653/v1/2023.emnlp-main.825},
  doi          = {10.18653/V1/2023.EMNLP-MAIN.825},
  bibsource    = {dblp computer science bibliography, https://dblp.org}
}

@inproceedings{llm-lingua-2,
  author       = {Zhuoshi Pan and
                  Qianhui Wu and
                  Huiqiang Jiang and
                  Menglin Xia and
                  Xufang Luo and
                  Jue Zhang and
                  Qingwei Lin and
                  Victor R{\"{u}}hle and
                  Yuqing Yang and
                  Chin{-}Yew Lin and
                  H. Vicky Zhao and
                  Lili Qiu and
                  Dongmei Zhang},
  editor       = {Lun{-}Wei Ku and
                  Andre Martins and
                  Vivek Srikumar},
  title        = {LLMLingua-2: Data Distillation for Efficient and Faithful Task-Agnostic
                  Prompt Compression},
  booktitle    = {Findings of the Association for Computational Linguistics, {ACL} 2024,
                  Bangkok, Thailand and virtual meeting, August 11-16, 2024},
  pages        = {963--981},
  publisher    = {Association for Computational Linguistics},
  year         = {2024},
  url          = {https://doi.org/10.18653/v1/2024.findings-acl.57},
  doi          = {10.18653/V1/2024.FINDINGS-ACL.57},
  bibsource    = {dblp computer science bibliography, https://dblp.org}
}

@inproceedings{dac,
  author       = {Yi Zhao and
                  Zuchao Li and
                  Hai Zhao and
                  Baoyuan Qi and
                  Liu Guoming},
  editor       = {Wanxiang Che and
                  Joyce Nabende and
                  Ekaterina Shutova and
                  Mohammad Taher Pilehvar},
  title        = {{DAC:} {A} Dynamic Attention-aware Approach for Task-Agnostic Prompt
                  Compression},
  booktitle    = {Proceedings of the 63rd Annual Meeting of the Association for Computational
                  Linguistics (Volume 1: Long Papers), {ACL} 2025, Vienna, Austria,
                  July 27 - August 1, 2025},
  pages        = {19395--19407},
  publisher    = {Association for Computational Linguistics},
  year         = {2025},
  url          = {https://aclanthology.org/2025.acl-long.952/},
  bibsource    = {dblp computer science bibliography, https://dblp.org}
}

@inproceedings{pyramidinfer,
  author       = {Dongjie Yang and
                  Xiaodong Han and
                  Yan Gao and
                  Yao Hu and
                  Shilin Zhang and
                  Hai Zhao},
  editor       = {Lun{-}Wei Ku and
                  Andre Martins and
                  Vivek Srikumar},
  title        = {PyramidInfer: Pyramid {KV} Cache Compression for High-throughput {LLM}
                  Inference},
  booktitle    = {Findings of the Association for Computational Linguistics, {ACL} 2024,
                  Bangkok, Thailand and virtual meeting, August 11-16, 2024},
  pages        = {3258--3270},
  publisher    = {Association for Computational Linguistics},
  year         = {2024},
  url          = {https://doi.org/10.18653/v1/2024.findings-acl.195},
  doi          = {10.18653/V1/2024.FINDINGS-ACL.195},
  bibsource    = {dblp computer science bibliography, https://dblp.org}
}

@inproceedings{fastv,
  author       = {Liang Chen and
                  Haozhe Zhao and
                  Tianyu Liu and
                  Shuai Bai and
                  Junyang Lin and
                  Chang Zhou and
                  Baobao Chang},
  editor       = {Ales Leonardis and
                  Elisa Ricci and
                  Stefan Roth and
                  Olga Russakovsky and
                  Torsten Sattler and
                  G{\"{u}}l Varol},
  title        = {An Image is Worth 1/2 Tokens After Layer 2: Plug-and-Play Inference
                  Acceleration for Large Vision-Language Models},
  booktitle    = {Computer Vision - {ECCV} 2024 - 18th European Conference, Milan, Italy,
                  September 29-October 4, 2024, Proceedings, Part {LXXXI}},
  series       = {Lecture Notes in Computer Science},
  volume       = {15139},
  pages        = {19--35},
  publisher    = {Springer},
  year         = {2024},
  url          = {https://doi.org/10.1007/978-3-031-73004-7\_2},
  doi          = {10.1007/978-3-031-73004-7\_2},
  bibsource    = {dblp computer science bibliography, https://dblp.org}
}

@article{dart,
  author       = {Zichen Wen and
                  Yifeng Gao and
                  Shaobo Wang and
                  Junyuan Zhang and
                  Qintong Zhang and
                  Weijia Li and
                  Conghui He and
                  Linfeng Zhang},
  title        = {Stop Looking for Important Tokens in Multimodal Language Models: Duplication
                  Matters More},
  journal      = {CoRR},
  volume       = {abs/2502.11494},
  year         = {2025},
  url          = {https://doi.org/10.48550/arXiv.2502.11494},
  doi          = {10.48550/ARXIV.2502.11494},
  eprinttype    = {arXiv},
  eprint       = {2502.11494},
  bibsource    = {dblp computer science bibliography, https://dblp.org}
}

@inproceedings{minference,
  author       = {Huiqiang Jiang and
                  Yucheng Li and
                  Chengruidong Zhang and
                  Qianhui Wu and
                  Xufang Luo and
                  Surin Ahn and
                  Zhenhua Han and
                  Amir H. Abdi and
                  Dongsheng Li and
                  Chin{-}Yew Lin and
                  Yuqing Yang and
                  Lili Qiu},
  editor       = {Amir Globersons and
                  Lester Mackey and
                  Danielle Belgrave and
                  Angela Fan and
                  Ulrich Paquet and
                  Jakub M. Tomczak and
                  Cheng Zhang},
  title        = {MInference 1.0: Accelerating Pre-filling for Long-Context LLMs via
                  Dynamic Sparse Attention},
  booktitle    = {Advances in Neural Information Processing Systems 38: Annual Conference
                  on Neural Information Processing Systems 2024, NeurIPS 2024, Vancouver,
                  BC, Canada, December 10 - 15, 2024},
  year         = {2024},
  url          = {http://papers.nips.cc/paper\_files/paper/2024/hash/5dfbe6f5671e82c76841ba687a8a9ecb-Abstract-Conference.html},
  bibsource    = {dblp computer science bibliography, https://dblp.org}
}

@inproceedings{flexprefill,
  author       = {Xunhao Lai and
                  Jianqiao Lu and
                  Yao Luo and
                  Yiyuan Ma and
                  Xun Zhou},
  title        = {FlexPrefill: {A} Context-Aware Sparse Attention Mechanism for Efficient
                  Long-Sequence Inference},
  booktitle    = {The Thirteenth International Conference on Learning Representations,
                  {ICLR} 2025, Singapore, April 24-28, 2025},
  publisher    = {OpenReview.net},
  year         = {2025},
  url          = {https://openreview.net/forum?id=OfjIlbelrT},
  bibsource    = {dblp computer science bibliography, https://dblp.org}
}

@inproceedings{mminference,
  author       = {Yucheng Li and
                  Huiqiang Jiang and
                  Chengruidong Zhang and
                  Qianhui Wu and
                  Xufang Luo and
                  Surin Ahn and
                  Amir H. Abdi and
                  Dongsheng Li and
                  Jianfeng Gao and
                  Yuqing Yang and
                  Lili Qiu},
  title        = {MMInference: Accelerating Pre-filling for Long-Context Visual Language
                  Models via Modality-Aware Permutation Sparse Attention},
  booktitle    = {Forty-second International Conference on Machine Learning, {ICML}
                  2025, Vancouver, BC, Canada, July 13-19, 2025},
  publisher    = {OpenReview.net},
  year         = {2025},
  url          = {https://openreview.net/forum?id=me6PfbATWM},
  bibsource    = {dblp computer science bibliography, https://dblp.org}
}

@inproceedings{quest,
  author       = {Jiaming Tang and
                  Yilong Zhao and
                  Kan Zhu and
                  Guangxuan Xiao and
                  Baris Kasikci and
                  Song Han},
  title        = {{QUEST:} Query-Aware Sparsity for Efficient Long-Context {LLM} Inference},
  booktitle    = {Forty-first International Conference on Machine Learning, {ICML} 2024,
                  Vienna, Austria, July 21-27, 2024},
  publisher    = {OpenReview.net},
  year         = {2024},
  url          = {https://openreview.net/forum?id=KzACYw0MTV},
  bibsource    = {dblp computer science bibliography, https://dblp.org}
}

@article{pqcache,
  author       = {Hailin Zhang and
                  Xiaodong Ji and
                  Yilin Chen and
                  Fangcheng Fu and
                  Xupeng Miao and
                  Xiaonan Nie and
                  Weipeng Chen and
                  Bin Cui},
  title        = {PQCache: Product Quantization-based KVCache for Long Context {LLM}
                  Inference},
  journal      = {Proc. {ACM} Manag. Data},
  volume       = {3},
  number       = {3},
  pages        = {201:1--201:30},
  year         = {2025},
  url          = {https://doi.org/10.1145/3725338},
  doi          = {10.1145/3725338},
  bibsource    = {dblp computer science bibliography, https://dblp.org}
}

@inproceedings{magicpig,
  author       = {Zhuoming Chen and
                  Ranajoy Sadhukhan and
                  Zihao Ye and
                  Yang Zhou and
                  Jianyu Zhang and
                  Niklas Nolte and
                  Yuandong Tian and
                  Matthijs Douze and
                  L{\'{e}}on Bottou and
                  Zhihao Jia and
                  Beidi Chen},
  title        = {MagicPIG: {LSH} Sampling for Efficient {LLM} Generation},
  booktitle    = {The Thirteenth International Conference on Learning Representations,
                  {ICLR} 2025, Singapore, April 24-28, 2025},
  publisher    = {OpenReview.net},
  year         = {2025},
  url          = {https://openreview.net/forum?id=ALzTQUgW8a},
  bibsource    = {dblp computer science bibliography, https://dblp.org}
}

@inproceedings{limitations-of-efim,
  author       = {Frederik Kunstner and
                  Philipp Hennig and
                  Lukas Balles},
  editor       = {Hanna M. Wallach and
                  Hugo Larochelle and
                  Alina Beygelzimer and
                  Florence d'Alch{\'{e}}{-}Buc and
                  Emily B. Fox and
                  Roman Garnett},
  title        = {Limitations of the empirical Fisher approximation for natural gradient
                  descent},
  booktitle    = {Advances in Neural Information Processing Systems 32: Annual Conference
                  on Neural Information Processing Systems 2019, NeurIPS 2019, December
                  8-14, 2019, Vancouver, BC, Canada},
  pages        = {4158--4169},
  year         = {2019},
  url          = {https://proceedings.neurips.cc/paper/2019/hash/46a558d97954d0692411c861cf78ef79-Abstract.html},
  bibsource    = {dblp computer science bibliography, https://dblp.org}
}

@inproceedings{distserve,
  author       = {Yinmin Zhong and
                  Shengyu Liu and
                  Junda Chen and
                  Jianbo Hu and
                  Yibo Zhu and
                  Xuanzhe Liu and
                  Xin Jin and
                  Hao Zhang},
  editor       = {Ada Gavrilovska and
                  Douglas B. Terry},
  title        = {DistServe: Disaggregating Prefill and Decoding for Goodput-optimized
                  Large Language Model Serving},
  booktitle    = {18th {USENIX} Symposium on Operating Systems Design and Implementation,
                  {OSDI} 2024, Santa Clara, CA, USA, July 10-12, 2024},
  pages        = {193--210},
  publisher    = {{USENIX} Association},
  year         = {2024},
  url          = {https://www.usenix.org/conference/osdi24/presentation/zhong-yinmin},
  bibsource    = {dblp computer science bibliography, https://dblp.org}
}

@inproceedings{splitwise,
  author       = {Pratyush Patel and
                  Esha Choukse and
                  Chaojie Zhang and
                  Aashaka Shah and
                  {\'{I}}{\~{n}}igo Goiri and
                  Saeed Maleki and
                  Ricardo Bianchini},
  title        = {Splitwise: Efficient Generative {LLM} Inference Using Phase Splitting},
  booktitle    = {51st {ACM/IEEE} Annual International Symposium on Computer Architecture,
                  {ISCA} 2024, Buenos Aires, Argentina, June 29 - July 3, 2024},
  pages        = {118--132},
  publisher    = {{IEEE}},
  year         = {2024},
  url          = {https://doi.org/10.1109/ISCA59077.2024.00019},
  doi          = {10.1109/ISCA59077.2024.00019},
  bibsource    = {dblp computer science bibliography, https://dblp.org}
}

@inproceedings{attention,
  author       = {Ashish Vaswani and
                  Noam Shazeer and
                  Niki Parmar and
                  Jakob Uszkoreit and
                  Llion Jones and
                  Aidan N. Gomez and
                  Lukasz Kaiser and
                  Illia Polosukhin},
  editor       = {Isabelle Guyon and
                  Ulrike von Luxburg and
                  Samy Bengio and
                  Hanna M. Wallach and
                  Rob Fergus and
                  S. V. N. Vishwanathan and
                  Roman Garnett},
  title        = {Attention is All you Need},
  booktitle    = {Advances in Neural Information Processing Systems 30: Annual Conference
                  on Neural Information Processing Systems 2017, December 4-9, 2017,
                  Long Beach, CA, {USA}},
  pages        = {5998--6008},
  year         = {2017},
  url          = {https://proceedings.neurips.cc/paper/2017/hash/3f5ee243547dee91fbd053c1c4a845aa-Abstract.html},
  bibsource    = {dblp computer science bibliography, https://dblp.org}
}

@inproceedings{pytorch,
  author       = {Adam Paszke and
                  Sam Gross and
                  Francisco Massa and
                  Adam Lerer and
                  James Bradbury and
                  Gregory Chanan and
                  Trevor Killeen and
                  Zeming Lin and
                  Natalia Gimelshein and
                  Luca Antiga and
                  Alban Desmaison and
                  Andreas K{\"{o}}pf and
                  Edward Z. Yang and
                  Zachary DeVito and
                  Martin Raison and
                  Alykhan Tejani and
                  Sasank Chilamkurthy and
                  Benoit Steiner and
                  Lu Fang and
                  Junjie Bai and
                  Soumith Chintala},
  editor       = {Hanna M. Wallach and
                  Hugo Larochelle and
                  Alina Beygelzimer and
                  Florence d'Alch{\'{e}}{-}Buc and
                  Emily B. Fox and
                  Roman Garnett},
  title        = {PyTorch: An Imperative Style, High-Performance Deep Learning Library},
  booktitle    = {Advances in Neural Information Processing Systems 32: Annual Conference
                  on Neural Information Processing Systems 2019, NeurIPS 2019, December
                  8-14, 2019, Vancouver, BC, Canada},
  pages        = {8024--8035},
  year         = {2019},
  url          = {https://proceedings.neurips.cc/paper/2019/hash/bdbca288fee7f92f2bfa9f7012727740-Abstract.html},
  bibsource    = {dblp computer science bibliography, https://dblp.org}
}

\appendix

\newpage

\section{Robustness to Representation Mismatch}
\label{sec:representation-mismatch}

A potential concern regarding POP is the representation mismatch introduced by layer pruning. 
In the prefill stage, bypassing a layer $l$ implies that the subsequent layer $l+1$ receives the input $x_l$ directly, rather than the expected $x_{l+1}$. 
Since the deep layers were trained to process specific feature distributions, one might expect this mismatch to accumulate, corrupting the KV cache and leading to catastrophic collapse in the decode stage.

However, our approach addresses this risk through both theoretical safeguards and empirical verification:

\noindent \textbf{Theoretical Safeguards via Virtual Gates.}
Theoretically, our importance estimation metric $\tilde{I}_l$ implicitly accounts for the sensitivity to representation mismatch. 
The score is derived from the gradient of the loss with respect to the virtual gate $g_l$:
\begin{equation*}
    \tilde{I}_l = \mathbb{E}\left[\left(\frac{\partial \mathcal{L}}{\partial g_l}\right)^2\right]
\end{equation*}
This gradient quantifies how much the final prediction loss $\mathcal{L}$ changes when layer $l$ is removed.
If skipping layer $l$ leads to a severe distortion in subsequent layers, the gradient would exhibit large variance, resulting in a high importance score.
Consequently, such layers would be retained by our strategy.
 
\noindent \textbf{Empirical Verification with Functional Robustness.}
To understand the physical mechanism of this robustness, we conduct a layer-wise analysis on Qwen3-VL using the WizardLM-V2-196K dataset. 
Specifically, we measure the internal consistency between the pruned and full models within the deep 1/3 layers (layer 25-36). 
We track the cosine similarity for 3 key representations: (1) the hidden states of each layer; (2) the KV cache of input tokens; and (3) the attention outputs of the decode stage.
As illustrated in Figure~\ref{fig:robustness}, we observe a striking contrast between representation drift and functional stability:

\begin{figure}[t]
    \centering
    \includegraphics[width=\linewidth]{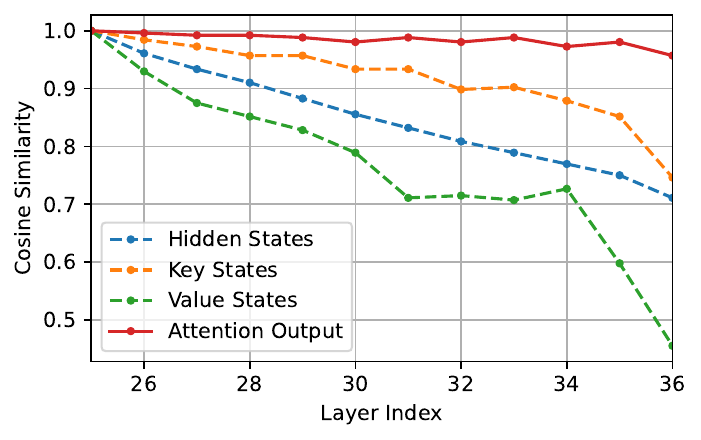}
    \caption{
        Cosine similarity of internal states between the POP-pruned model and the full model across deep layers (25-36). 
    }
    \label{fig:robustness}
\end{figure}

\begin{itemize}[nosep]
    \item \textbf{Representation Drift:}
    As expected, skipping layers accumulates numerical deviations. 
    The similarity of hidden states gradually drops from 1.0 down to 0.71 (blue dashed line), and the value states in the KV cache (green dashed line) show even greater divergence, dropping to as low as 0.46. 
    This confirms that the vector space of the pruned model indeed drifts from the original trajectory.
    \item \textbf{Functional Stability:}
     Although the attention module receives drifted keys and values as inputs, its output maintains high fidelity.
     Specifically, the attention output (red solid line) consistently maintains a high cosine similarity, staying above 0.96 across all measured layers (25-36), significantly outperforming other internal states.
     This indicates that the attention mechanism acts as a robust stabilizer: the weighted aggregation over the context window effectively smooths out the noise from individual drifted tokens, ensuring the semantic information passed to the next layer remains valid.
\end{itemize}

\noindent \textbf{Conclusion.} 
Our analysis reveals that POP succeeds because the deep layers possess intrinsic functional robustness, where the attention mechanism compensates for the representation drift.
Our virtual gate mechanism correctly captures this property: the low gradient variance calculated for these deep layers implies that the loss landscape is insensitive to the observed representation mismatch.

\section{Integration with Orthogonal Methods}

\begin{table*}[htb]
\centering
\caption{
Accuracies on long context tasks and TTFT speedup ratios. 
Experiments are conducted with greedy decoding (Temperature = 0). 
TTFT speedup is measured with batch size 1, input length 32K.
}
\label{tab:orthogonality}
\small
\begin{tabular}{lccc}
\toprule
\textbf{Method} & \textbf{TTFT Speedup} & \textbf{HotpotQA} & \textbf{MultiFieldQA} \\
\midrule
Full Model                                      & 1.00$\times$ & 57.83 & 56.00 \\
\midrule
FlexPrefill ($\gamma = 0.95$)                   & 1.13$\times$ & 56.39 & 54.41 \\
FlexPrefill ($\gamma = 0.8$)                    & 1.22$\times$ & 47.79 & 48.55 \\
FlexPrefill ($\gamma = 0.95$) + POP (31.25\%)   & \textbf{1.54$\times$} & \textbf{54.34} & \textbf{53.42} \\
\midrule
LLMLingua-2 (0.4)                               & 1.37$\times$ & 56.11 & 42.49 \\
LLMLingua-2 (0.3)                               & 1.53$\times$ & 50.92 & 39.90 \\
LLMLingua-2 (0.4) + POP (31.25\%)               & \textbf{1.56$\times$} & \textbf{51.74} & \textbf{42.31} \\
\bottomrule
\end{tabular}
\end{table*}

As noted in our Related Work section (Section~\ref{sec:related-work}), POP (model pruning) is orthogonal to both sparse attention and token pruning methods. 
Technically, while token pruning reduces the sequence length $N$ (temporal dimension) 
and sparse attention reduces computation within attention blocks (computational sparsity), 
POP physically removes entire transformer layers $L$ (depth dimension). 
Thus, POP reduces the computational cost per token regardless of the sequence length or attention pattern, making it naturally complementary to these approaches.

To demonstrate this, we integrate POP with FlexPrefill~\cite{flexprefill} and LLMLingua-2~\cite{llm-lingua-2}. 
Considering that both baselines are primarily designed to handle long-context challenges, we conducted additional experiments on Llama-3.1-8B-Instruct using the HotpotQA and MultiFieldQA datasets. 
Experimental results are shown in Table~\ref{tab:orthogonality}.

Experimental results demonstrate that \textbf{POP exhibits strong orthogonality with both sparse attention and token pruning methods}. 
Specifically, aggressively increasing sparsity in baselines (e.g., FlexPrefill $\gamma$ = 0.8) yields limited speedup ($1.22\times$) but suffers from significant accuracy drops (57.83 $\to$ 47.79 on HotpotQA). 
In contrast, combining a conservative setting of these methods (e.g., FlexPrefill $\gamma$ = 0.95 or LLMLingua-2 ratio = 0.4) with POP achieves significantly higher speedups (${\sim}$1.56$\times$) while preserving higher accuracies (54.34 / 51.74 on HotpotQA). 
This confirms that the joint application of POP with these orthogonal methods yields a superior Pareto frontier in terms of accuracy and efficiency.

\end{document}